\documentclass[a4paper,fleqn]{cas-dc}

\usepackage[authoryear,longnamesfirst]{natbib}

\def\tsc#1{\csdef{#1}{\textsc{\lowercase{#1}}\xspace}}
\tsc{WGM}
\tsc{QE}
\tsc{EP}
\tsc{PMS}
\tsc{BEC}
\tsc{DE}

\begin{document}
\let\WriteBookmarks\relax
\def\floatpagepagefraction{1}
\def\textpagefraction{.001}



\title [mode = title]{EPANet: Efficient Path Aggregation Network for Underwater Fish Detection}                      



%
\author[1,2,3,4]{Jinsong Yang}





\affiliation[1]{organization={Dalian Key Laboratory of Smart Fishery},
    city={Dalian},
    country={China}}

\author[1,2,3,4]{Zeyuan Hu}[]
\cormark[1]
\ead{huzy@dlou.edu.cn}
\author[5]{Yichen Li}[]

\affiliation[2]{organization={College of Information Engineering, Dalian Ocean University},
    country={China}}

\affiliation[3]{organization={Liaoning Provincial Key of Marine Information Technology},
    country={China}}
    
\affiliation[4]{organization={Key Laboratory of Environment Controlled Aquaculture},
    city={Dalian},
    country={China}}

\affiliation[5]{organization={Dalian Polytechnic University School of International Education},
    city={Dalian},
    country={China}}
\cortext[cor1]{Corresponding author}

\fntext[fn1]{0000-0002-3673-356X}

\begin{abstract}
Underwater fish detection (UFD) remains a challenging task in computer vision due to low object resolution, significant background interference, and high visual similarity between targets and surroundings. Existing approaches primarily focus on local feature enhancement or incorporate complex attention mechanisms to highlight small objects, often at the cost of increased model complexity and reduced efficiency. To address these limitations, we propose an efficient path aggregation network (EPANet), which leverages complementary feature integration to achieve accurate and lightweight UFD. EPANet consists of two key components: an efficient path aggregation feature pyramid network (EPA-FPN) and a multi-scale diverse-division short path bottleneck (MS-DDSP bottleneck). The EPA-FPN introduces long-range skip connections across disparate scales to improve semantic-spatial complementarity, while cross-layer fusion paths are adopted to enhance feature integration efficiency. The MS-DDSP bottleneck extends the conventional bottleneck structure by introducing finer-grained feature division and diverse convolutional operations, thereby increasing local feature diversity and representation capacity. Extensive experiments on benchmark UFD datasets demonstrate that EPANet outperforms state-of-the-art methods in terms of detection accuracy and inference speed, while maintaining comparable or even lower parameter complexity.
\end{abstract}


\begin{highlights}
\item We propose EPA-FPN, a lightweight yet effective
FPN that improves semantic-spatial complementarity
via long-range and lateral connections while reducing
path redundancy.
\item We introduce MS-DDSP bottleneck, a novel module
that performs fine-grained feature division and em-
ploys diverse convolutions to increase the richness and
scale-awareness of local features.
\item We develop EPANet, a complete UFD detection frame-
work based on EPA-FPN and MS-DDSP bottleneck,
which outperforms state-of-the-art methods such as
YOLOv11s in both accuracy and inference speed
while maintaining a similar parameter budget.
\end{highlights}


\maketitle

\section{introduction}

Underwater fish detection (UFD), a critical branch of object detection, has gained increasing attention due to its broad applications in marine ecological monitoring, biodiversity assessment, and intelligent fisheries~\cite{WANG20231,CAI2025130087,WANG2022104458,XU2023204,ZHOU2024102680,10852283}. Despite the progress in deep learning-based detection frameworks, UFD remains particularly challenging due to the low resolution of underwater targets, visual similarity between fish and background, and environmental factors such as low lighting and water turbidity~\cite{JIAN2021116088,FU2023243,TENG2024117958}. These issues often lead to inaccurate localization and a high rate of missed or false detections.

Recent studies in aerial and remote sensing detection have shown that incorporating frequency-domain interactions and refined pyramid structures can significantly improve detection under low-resolution or noisy environments~\cite{weng2024enhancing,li2024lr,qiao2022novel,weng2023novel}. Inspired by these findings, we argue that enhancing feature complementarity and increasing local feature diversity—rather than solely deepening the network or stacking attention modules—offers a more efficient pathway toward robust UFD.
Driven by the success of deep convolutional neural networks (DCNNs), methods such as YOLO-Fish~\cite{MUKSIT2022101847} and DeepFish~\cite{QIN201649} have been proposed to enhance feature extraction capabilities in underwater scenarios. However, these approaches typically increase model complexity, while ignoring key challenges such as spatial information degradation and inefficient feature fusion—both of which limit performance under constrained underwater conditions~\cite{FENG2024102758,YU2023102108}.

\begin{figure}[t]
    \centering
    \includegraphics[width=1\linewidth]{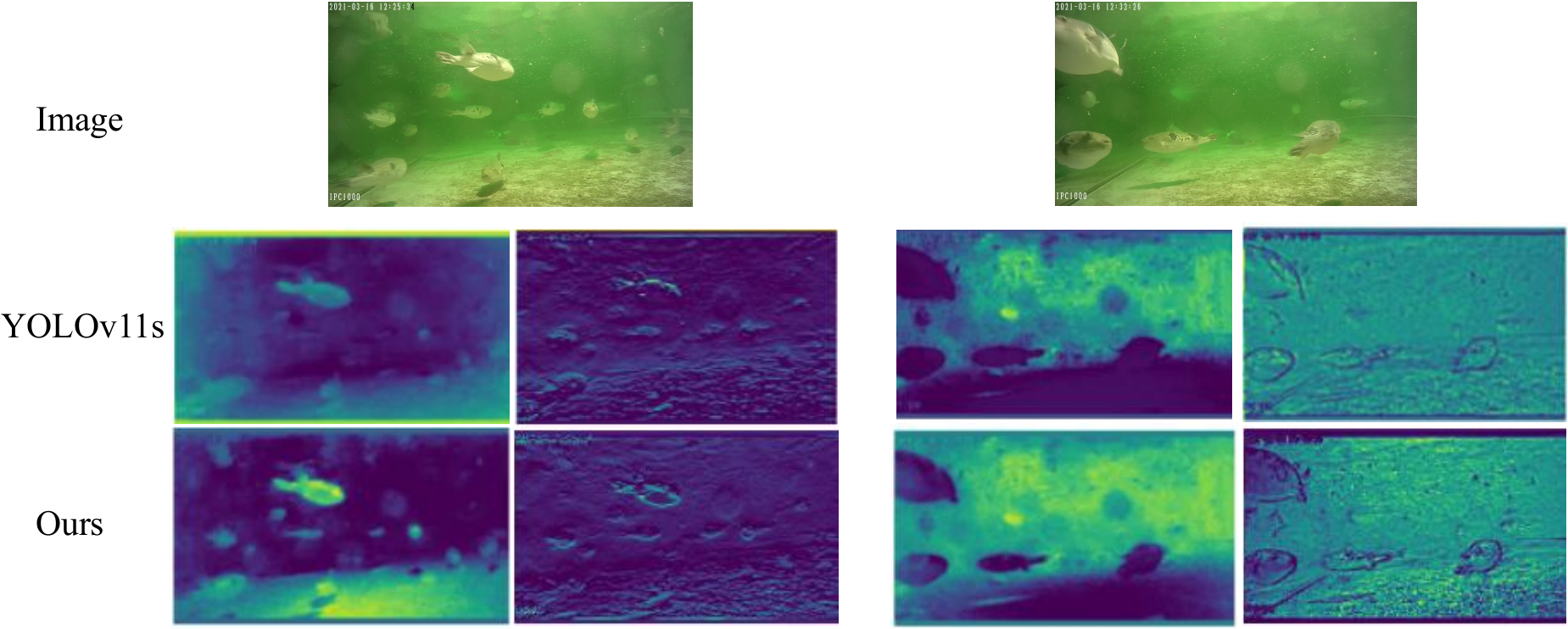}
\caption{Feature map visualization comparison. Our method can better distinguish between objects and their background, allowing for more accurate extraction of local detail features.}
\label{fig: Feature map}
\end{figure}

To address these challenges, researchers have turned to feature pyramid networks (FPNs)\cite{ZHU2022127,HAN2024127809,MENG2025129196} and feature fusion strategies\cite{LI2025130102,LIN2025130351,AHMAD2025129725} to enhance multi-scale feature representation and semantic-spatial alignment. Representative works such as PANet~\cite{2018arXiv180301534L}, Bi-FPN~\cite{9156454}, and NAS-FPN~\cite{8954436} achieve improved performance through bottom-up and top-down information pathways. Likewise, attention-guided or progressive fusion modules~\cite{Dai_2021_WACV,LI2025130755,ZHU2025128898} aim to refine local details while preserving global context. However, most of these techniques were initially designed for natural images or aerial scenes and often fail to generalize effectively to underwater domains.

In this work, we propose an efficient and lightweight architecture named EPANet, tailored specifically for underwater fish detection. EPANet integrates two key components: an efficient path aggregation feature pyramid network (EPA-FPN) and a multi-scale diverse-division short path bottleneck (MS-DDSP bottleneck). EPA-FPN improves multi-scale semantic-spatial fusion via long-range and cross-layer connections, while MS-DDSP bottleneck introduces diverse convolutional branches and fine-grained decomposition to enrich local feature representation. As shown in Fig.~\ref{fig: Feature map}, our method yields clearer boundaries and stronger discrimination between objects and background, especially under blurred or low-light conditions.

To summarize, our contributions are three-fold:
\begin{itemize}
\item We propose EPA-FPN, a lightweight yet effective FPN that improves semantic-spatial complementarity via long-range and lateral connections while reducing path redundancy.
\item We introduce MS-DDSP bottleneck, a novel module that performs fine-grained feature division and employs diverse convolutions to increase the richness and scale-awareness of local features.

\item We develop EPANet, a complete UFD detection framework based on EPA-FPN and MS-DDSP bottleneck, which outperforms state-of-the-art methods such as YOLOv11s in both accuracy and inference speed while maintaining a similar parameter budget.
\end{itemize}

\section{related work}

\subsection{underwater fish detection methods}

Underwater fish detection (UFD) methods can be grouped into two-stage and one-stage paradigms.
Two-stage detectors, based on the RCNN framework~\cite{7410526,SUN201842}, generate region proposals via an RPN followed by classification and regression. iFaster RCNN~\cite{10038226} integrates FPN and DIoU loss for better multi-scale localization, while CAM-RCNN~\cite{YI2024127488} introduces CoordConv and group normalization to enhance spatial awareness and generalization. However, these methods incur high computational cost due to extensive proposal filtering, limiting real-time deployment.
One-stage detectors directly regress bounding boxes from dense grids, offering greater speed. YOLO-Fish~\cite{MUKSIT2022101847} enhances small-object detection using SPP and optimized upsampling. RC-YOLOv5~\cite{YI2024127488} combines Res2Net and coordinate attention to improve robustness in complex underwater environments. SUR-Net~\cite{9851605} jointly handles detection and segmentation with an attention-enhanced U-Net and adaptive augmentation, boosting performance with limited data.

Despite these advances, many methods trade off between accuracy and efficiency, relying on stacked modules or complex backbones. Inspired by recent progress in efficient conditional generation~\cite{shen2024imagpose,shen2025imagdressing}, where structured priors and feature decomposition yield expressive representations with low overhead, we redesign the FPN to enhance spatial complementarity and feature diversity. Our approach avoids excessive depth or attention layers, achieving strong UFD performance while maintaining high inference efficiency.

\begin{figure*}[t]
	    \centering
        \includegraphics[width=0.99\linewidth]{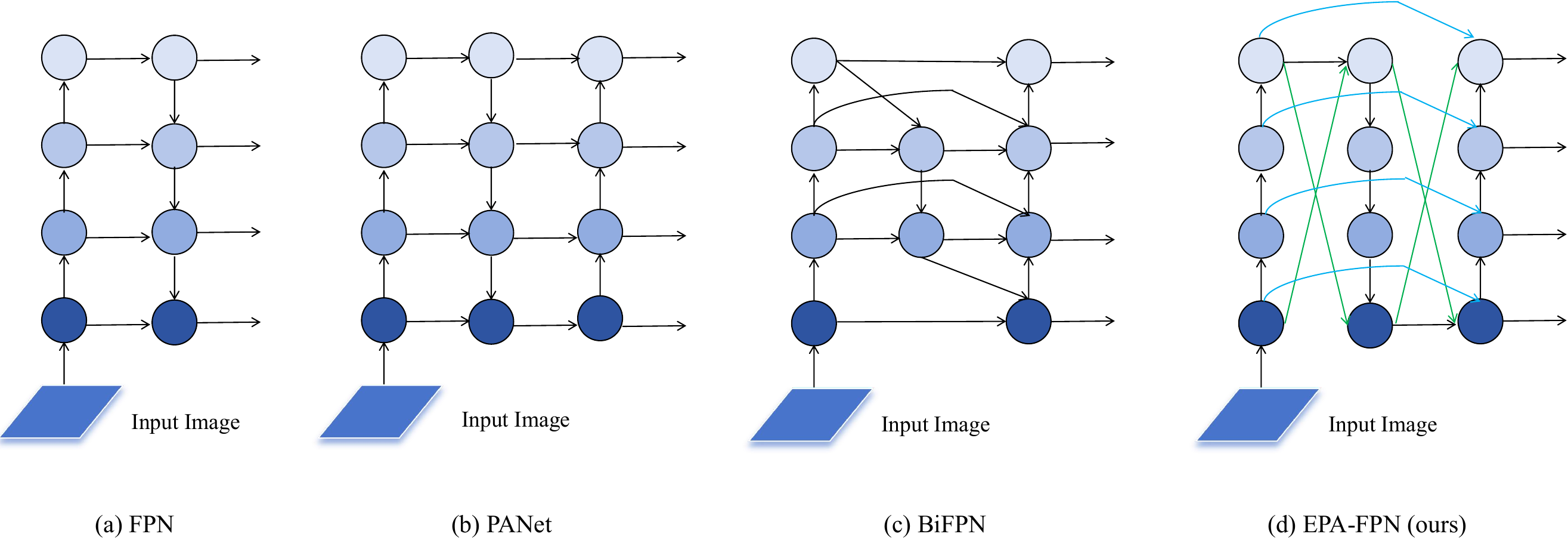}
        \caption{\textbf{Feature Pyramid Network design --}
        	(a) FPN introduces a top-down pathway to fuse multi-scale features; (b) PANet adds an additional bottom-up pathway on top of FPN; (c) BiFPN trades off accuracy and efficiency through horizontal cross-connection and compound scaling; (d) is our EPA-FPN that further optimizes feature fusion and information complementarity.
            }
       \vskip -0.1in
        \label{fig:epafpn}                                              
\end{figure*} 

\subsection{efficient fpn design}

Feature pyramid networks (FPNs) play a key role in multi-scale object detection by integrating semantic and spatial cues across different feature levels~\cite{JIE2021107,XIE2025128775,QUYEN2023104}. Dy-FPN~\cite{2020arXiv201200779Z} introduces dynamic branch selection with gating mechanisms to improve inference adaptability. NAS-FPN~\cite{8954436} uses neural architecture search to optimize cross-scale fusion, while Bi-FPN~\cite{9156454} employs learnable weights to balance feature contributions. PANet~\cite{2018arXiv180301534L} enhances spatial context aggregation through path-level fusion, improving robustness in dense detection tasks.
However, most existing designs prioritize structural diversity and fusion depth, while overlooking spatial degradation in deep layers. Although PANet considers spatial-semantic aggregation, its design introduces path redundancy and underutilizes spatial detail, limiting aggregation efficiency. To address this, we propose an efficient path aggregation FPN (EPA-FPN) that combines vertical long-range and horizontal cross-scale connections for enhanced spatial-semantic integration.

EPA-FPN maintains spatial precision by injecting low-level detail into high-level semantic features, which helps distinguish objects from complex backgrounds. This design is aligned with prior efforts on structure-aware representation learning, such as PB-SL~\cite{shen2023pbsl}, which introduces bipartite-aware similarity learning for cross-modal retrieval, and Triplet Contrastive Learning~\cite{shen2023triplet}, which focuses on robust feature separation under unsupervised settings. These studies highlight the importance of structure-aware fusion and adaptive representation, which we incorporate into EPA-FPN for compact yet discriminative multi-scale modeling in UFD.

\section{Proposed Method}

\subsection{Overview}
The overall structure of our EPANet is shown in Fig. \ref{fig:epanet}. Given an input image after Backbone feature extraction, we use the Efficient Path Aggregation Feature Pyramid Network (EPA-FPN) to achieve long-range spatial semantic information aggregation and cross-scale efficient multi-feature fusion, thereby enhancing the detection ability of underwater targets of different scales. In order to better distinguish targets from backgrounds, we designed the MS-DDSP Bottleneck module, which captures a wider receptive field and more diverse features by dividing the traditional Bottleneck more finely and extracting features using different convolution methods. Then, the c2f module (MS-c2f) containing the MS-DDSP Bottleneck module is added to the path with higher contribution to achieve further enhancement, and the corresponding detection head is named SP-Detect. Below, we will introduce these modules in detail.

\subsection{EPA-FPN}  
In this section, we first explain the problem of multi-scale feature fusion, and then introduce the main ideas of our proposed EPA-FPN: horizontal cross-scale connection and vertical large-scale difference long-range aggregation achieve more efficient feature fusion and better information complementarity, respectively.

Multi-scale feature fusion aims to aggregate features at different resolutions. Formally speaking, given a multi-scale feature list $C = (C_1, C_2,...)$, where $C_i$ represents the
feature at level $i$, our goal is to find a transformation f that can effectively aggregate different features and output a new feature list $P = f (C)$. As a concrete example, Figure \ref{fig:epafpn} (a) shows a traditional top-down FPN. Its input features can be expressed as $C = (C_2, C_3, C_4)$, where $C_i$ represents the feature level with a resolution of $1/2_i$ of the input image. For example, if the input resolution is 640x640, then $C_3$ represents feature level 3 with a resolution of 80x80 ($640/2^3=80$), and $C_5$ represents feature level 5 with a resolution of 20×20. The traditional FPN aggregates multi-scale features in a top-down manner:
\begin{align}
    P_4 &= {Conv}_{3\times3}\left( {Conv}_{1\times1}(C_4) + \uparrow(P_5) \right) \label{eq:p4} \\
    P_3 &= {Conv}_{3\times3}\left( {Conv}_{1\times1}(C_3) + \uparrow(P_4) \right) \label{eq:p3} \\
    P_2 &= {Conv}_{3\times3}\left( {Conv}_{1\times1}(C_2) + \uparrow(P_3) \right)
\end{align}
Where $\uparrow$ represents the upsampling operation for resolution matching, and  ${Conv}{1\times1}$ represents a 1x1 convolution operation for feature processing to adjust the number of channels.

Traditional top-down FPN is limited by unidirectional information flow, which makes it difficult to utilize the gradually weakened spatial information during network propagation. To solve this problem, PANet adds an additional bottom-up path aggregation network, as shown in Figure \ref{fig:epafpn} (b), which adds accurate spatial information to lower-level features while shortening the information path between lower-level features and top-level features. However, the spatial information added by PANet comes from feature maps of adjacent scales, while in fact, more significant spatial information often exists in higher-level features, and more significant semantic information exists in lower-level features. In other words, the greater the difference between the levels, the more obvious the difference between semantic and spatial information. Therefore, we think of aggregating the most significant spatial information at the top layer directly with the most significant semantic information at the lowest layer, which can not only ensure the best aggregation effect, but also make full use of spatial and semantic information to reduce information redundancy. At the same time, we also found that not all aggregation layers have the same impact on the output (we call this impact contribution). After verification, we found that the contribution of the middle layer is lower than that of the top and bottom layers. Therefore, we prune the paths with low contribution and only retain those with high contribution (as shown in the \textcolor[rgb]{0,0.5,0}{green line} in Figure \ref{fig:epafpn} (d)).

In underwater scenes, multiple objects are randomly distributed in different positions of the image, causing these objects to appear in different sizes. In addition, due to insufficient underwater light and turbid water, the object and the background are highly similar in feature dimensions such as color and texture, which greatly interferes with our detection. Although accurately locating the object position by aggregating spatial information can effectively solve the problem of missed object detection, it still has limitations in distinguishing the difference between the object and the background. The BiFPN structure (as shown in Figure \ref{fig:epafpn} (c)) achieves fast multi-scale feature fusion through horizontal cross-connection. Inspired by it, from the perspective of features, we also use a similar horizontal cross-connection structure to directly fuse shallow features containing more details with deep layers. The difference is that we use a single path of horizontal cross-connection to replace the original two paths, which can not only ensure the original information aggregation effect but also reduce nearly half of the path parameters, thereby effectively balancing the problem of increased computational complexity caused by shallow detail features.(as shown in the \textcolor[rgb]{0,0,0.9}{blue line} in Figure \ref{fig:epafpn} (d)).With these optimizations, we name the new feature network as efficient path aggregation feature pyramid network (EPA-FPN).

\begin{figure*}[t]
	    \centering
        \includegraphics[width=0.99\linewidth]{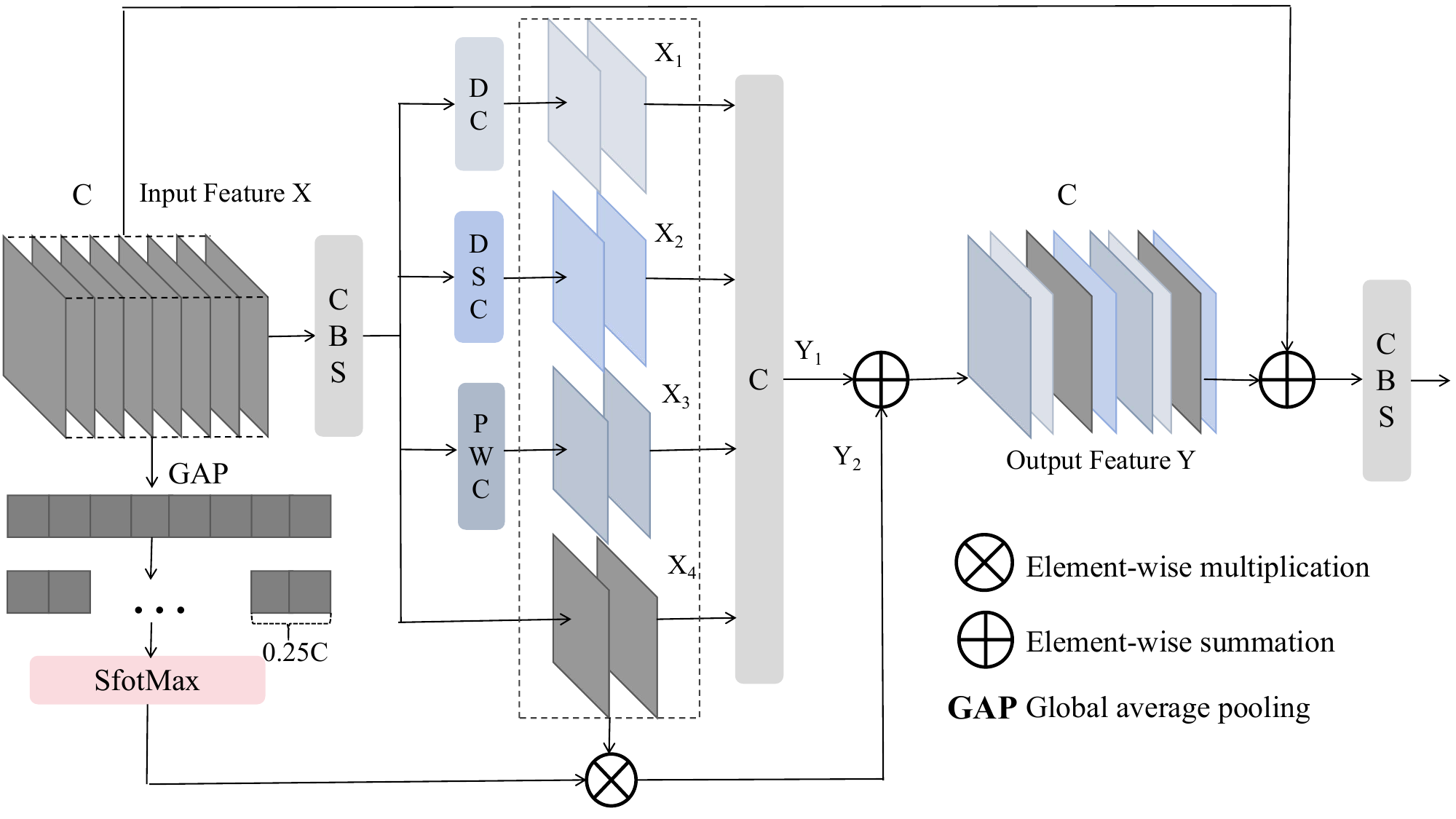}
        \caption{\textbf{Bottleneck network design --}
        	CBS represents the combination of convolution, batch normalization, and SiLu activation function. DC represents dilated convolution, DSC represents depthwise separable convolution, and PWC represents pointwise convolution.
            }
       \vskip -0.1in
        \label{fig:bottleneck}                                                  
\end{figure*} 

\subsection{MS-DDSP Bottleneck}  
In the previous section, we used horizontal cross-connections to directly fuse shallow detail features to solve the problem of high similarity between the object and the background. In order to better extract these local detail features, in this section, we designed a new BottleNeck network~\cite{10533618}~\cite{2024YOLOv10}~\cite{10533611} and named it MS-DDSP Bottleneck, and named its C2f module C2f-MS. Below, we will introduce  C2f and BottleNeck, as well as the main ideas of our MS-DDSP Bottleneck, respectively.

Cross stage partial Fusion bottleneck with 2 convolutions (C2f),as one of the key components of YOLOv8, has a unique dual-branch topology and feature decoupling strategy that shows superiority in network feature extraction efficiency and multi-scale modeling capabilities. When the input is $X$ and the output is $Y$, it is expressed as follows:

\begin{equation}
Y = {Conv}_{1 \times 1} \left( {Concat} \left( {Bottleneck}^{N}(F_a),\ F_b \right) \right) + X
\end{equation}
\begin{equation}
F_a,\ F_b = {Split}\left( {Conv}_{1 \times 1}(X) \right)
\end{equation}
Where ${Conv}{1\times1}$ represents a 1x1 convolution operation, ${Bottleneck}^{N}$ represents an N-times stacked Bottleneck module, ${Concat}$ represents a channel dimension concatenation operation, and ${Split}$ represents a channel averaging operation.

In particular, the main branch of C2f uses n Bottleneck units to achieve efficient extraction of deep semantic information. Specifically, each Bottleneck uses depthwise separable convolution (DSC) to extract features and uses SiLU activation function for nonlinear transformation. Finally, residual Skip Connection is used to retain the original input features to alleviate the vanishing gradient of the deep network:

\begin{equation}
{B}(F) = {Conv}_{3\times3} \circ {SiLU} \circ {BN} \circ {Conv}_{1\times1}(F) + F
\end{equation}
\begin{equation}
Y = {SiLU}\left( {Conv}_{1\times1}(F_{{concat}}) + X \right)
\end{equation}
Where $\circ$ represents the function composite operation,  
  $SiLU$ represents the SiLU activation function defined as $SiLU = x \cdot \sigma(x)$, and $F_{concat}$ represents the concatenated feature tensor.

To better extract local detail features to effectively distinguish the object from the background, we proposed the MS-DDSP Bottleneck. Different from the ordinary BottleNeck, we made a more detailed division inside the BottleNeck structure to capture more comprehensive detail features and obtain a wider multi-scale receptive field. As shown in Figure \ref{fig:bottleneck}, the input feature map $X\in \mathbb{R}^{(C\times H\times W)}$ is divided into four equal parts after adjustment by CBS, each of which is $X\in \mathbb{R}^{(0.25C\times H\times W)}$.
\begin{equation}
    {Split}_{C/4}(X) = [X_1, X_2, X_3, X_4] 
\end{equation}

Specifically, in the first part, we use staged dilated convolution (DC), and the dilation rate is set to 1, 2, and 4, respectively. In this way, local to global features are extracted in stages, and the progressive expansion of the receptive field can effectively solve the problem of different target sizes in underwater scenes. Finally, a feature map is obtained as the output $X_1$. 
\begin{equation}
    X_1 = {Upsample}  \Big[ \big( (X'_1 \ast_{d=1} W_1) \ast_{d=2} W_2 \big) \ast_{d=4} W_3 \Big] 
\end{equation}

In the second part, we use deep separable convolution (DSC). The channel-by-channel calculation characteristics of DSC can reduce the risk of overfitting background clutter, thereby effectively suppressing underwater background noise. Finally, a feature map is obtained as the output $X_2$.
\begin{equation}
    X_2 = W_{\text{pw}} \ast {(W_{\text{dw}} \ast X'_2)}
\end{equation}

We use pointwise convolution  (PWC) in the third part. The channel weighting mechanism of PWC can automatically learn to suppress low-value channels related to the underwater background and strengthen the response of target-related channels. Then, we obtain the output $X_3$. 
\begin{equation}
    X_3 = W_{\text{pwc}} \ast X'_3  \quad 
\end{equation}

To retain the detailed features to the greatest extent while ensuring the diversity of the feature maps, we do not perform any processing on the fourth part $X_4$. After concat, we finally get feature maps with different receptive fields, different detailed features, and different sizes.
\begin{equation}
    Y_1 = Concat\left( [X_1, X_2, X_3, X_4] \right) \quad 
\end{equation}

So far, we have obtained the feature map $Y_1$ after the concatenation of four refined features. Since these four features come from different output channels, a fusion method is needed to adaptively select more discriminative features. Specifically, we first use global average pooling (GAP) to collect global spatial and semantic information and represent it using channel statistics $S\in \mathbb{R}^{C\times1\times1}$,$n\in{\{1,2,3,4}\}$, which can be represented as:
\begin{equation}
S^n = {GAP}(X^n) = \frac{1}{H \times W} \sum_{i=0}^{H-1} \sum_{j=0}^{W-1} X_{i,j}^n
\end{equation}
Next, we use channel-wise soft attention operation to generate feature selectivity weights $\beta^n \in{\mathbb{R}^C} $, which are defined as follows:
\begin{equation}
 \beta^n = \frac{e^{S^n}}{\sum_{k=1}^{4} {e^{S^k}}}.
\end{equation}

Finally, under the guidance of the feature selection weight $\beta^n$, the feature map$Y_2$ is obtained by merging the branch outputs, and the final output $Y$ is obtained by adding $Y_1$ and $Y_2$, as shown below:
\begin{equation}
Y_2 = {Concat}\left( \left[ X_n \odot \beta^n \right] \right) 
\end{equation}
\begin{equation}
Y = Y_{1} + Y_{2}
\end{equation}
Where $\odot$ represents channel-by-channel multiplication, i.e., Hadamard product. $Y,Y_1 ,Y_2\in \mathbb{R}^{C \times H \times W}$,$n\in{\{1,2,3,4}\}$.

 \begin{figure*}[t]
	    \centering
        \includegraphics[width=0.99\linewidth]{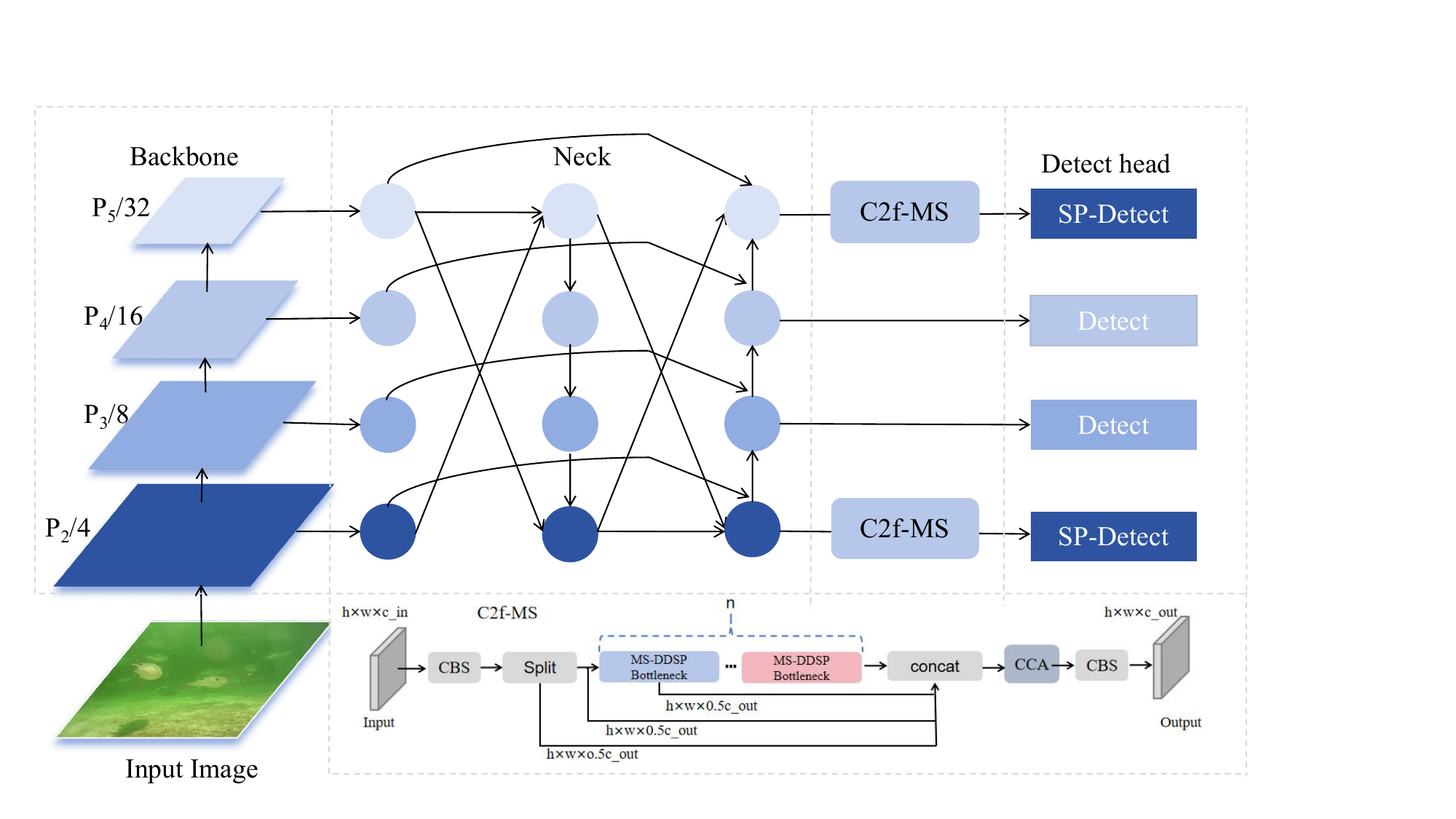}
        \caption{\textbf{EPA-Net architectur --}
        	 It uses CSPDarknet as the backbone network, EPA-FPN as the feature network and forms the neck network together with MS-c2f. The structure of the MS-c2f module is at the bottom of the figure.
            }
       \vskip -0.1in
        \label{fig:epanet}                                                  
\end{figure*} 

Overall, by dividing Bottleneck more finely and using dilated convolution (DC), depthwise separable convolution (DSC), and pointwise convolution (PWC) to extract more diverse features, and using attention operations to focus on important features, MS-DDSP Bottleneck is arranged according to the "divergence-extraction-reorganization" strategy, which can effectively capture more diverse features and richer information, thus helping the model to distinguish objects and backgrounds in underwater scenes better.

\section{Experiment and Analysis}
To validate the proposed method’s superiority, it is compared with multiple state-of-the-art (SoTA) approaches on three UFD datasets for different scenarios, namely, DeepFish, AquaFishSet, and UW-BlurredFish.

\subsection{Datasets}
\textbf{\emph{DeepFish}}
DeepFish~\cite{QIN201649} is one of the most well-known underwater fish scene datasets, which consists of approximately 40 thousand images collected underwater from 20 habitats in marine environments. We selected 6517 images suitable for the object detection task. These images are divided into sets with numbers of 4610, 500, and 1407 for training, validation, and testing. 

\textbf{\emph{AquaFishSet}}
AquaFishSet provides 1375 images of underwater close-up school fish scenes, which are partitioned into 367, 101, and 233 for training, validation, and test. The background of the image is complex, and the school of fish is blocked to a certain extent, which places relatively high demands on the performance of the real-time model.

\textbf{\emph{UW-BlurredFish}}
To further verify the effectiveness of our method, we also built a very challenging underwater fish dataset, UW-BlurredFish, for testing, which provides 600 blurred underwater fish images. This dataset has high background similarity and low target resolution, which is very challenging for real-time models.

\subsection{Evaluation Metrics}  
We mainly use Average Precision(AP) to evaluate the performance of the model, which is the most popular metric in various object detection tasks. It reflects both the precision and the recall ratio of the objects’ detection results.  Under the scenarios of the UFD task, we pay far more attention to whether the target can be found than to its location accuracy. Therefore, IoU = 0.5 naturally becomes the main evaluation criterion. At the same time, in order to further evaluate the false detection and missed detection of the model in underwater scenarios of high background similarity and low target resolution, we use precision and recall, respectively. Of course, we also use Params and FLOPs to evaluate model complexity. In addition, in order to evaluate the robustness of the model in complex scenarios and its adaptability to different requirements, we use the mean average precision(mAP) of the IOU threshold from 0.5 to 0.95, with a step size of 0.05, as an evaluation metric. In particular, there are very few positive samples in DeepFish, so we added the F1 score as an additional evaluation metric.

\begin{table*}[t]
    \centering
    \normalsize
    \caption{Comparison of EPA-Net and SoTA methods on \textbf{DeepFish} dataset. The best and second-best performances are highlighted in \textbf{bold} and \underline{underlined}, respectively.}
    \setlength{\tabcolsep}{6pt}
    \begin{tabular}{l|c|c|c|c|c|c|c c}
        \hline
        \textbf{Model} & \textbf{Param. (M) }& \textbf{FLOPs (G)}& \textbf{P}& \textbf{R} & \textbf{mAP$_{50}$ } & \textbf{mAP$_{50-95}$ } & \textbf{F1 score}\\
        \hline
        Faster R-CNN~\cite{7410526}
 &41.3 & 138.5 & 83.4 & 75.8 & 83.4 & 50.4 & 79.5\\
CAM-RCNN~\cite{YI2024127488}
 &44.0 & 189.0 & 86.2 & 77.4 & 84.8 & 51.4 & 81.5\\
 Mask R-CNN~\cite{8237584}
 &43.99 & 188.4 & 85.6 & 77.2 & 84.2 & 51.2 & 81.2\\
 YOLOv5n~\cite{10533615}
 &\textbf{1.92} & 4.5 & 86.2 & 79.6 & 86.8 & 51.5 & 82.8\\
 RC-YOLOv5~\cite{Li2023} 
 &\underline{2.2} & 6.3 & 87.2 & 76.4 & 87.2 & 51.9 & 81.4\\
  YOLOv8n~\cite{10533618}
 &3.01 & 8.9 & 85.3 & 83.1 & 88.1 & 51.5& 84.2\\
 YOLOv10n~\cite{2024YOLOv10}
 &2.3 & 6.7 & 87.9 & 82.6 & 88.6 & 51.3& 85.2\\
 YOLOv11n~\cite{10533611}
 &2.6 & 6.5 & 87.0 & 81.5 & 88.1 & 51.2& 84.2\\
 CEH-YOLO~\cite{FENG2024102758}
 &4.4 & 11.6 & 87.2 & 80.2 & 87.5 & 50.2& 83.6\\
 YOLO-Fish~\cite{MUKSIT2022101847} 
 &61.8 & 65.9 & 86.8 & 82.2 & 88.6 & 52.2& 84.1\\
 SUR-Net~\cite{9851605} 
 &30.2 & 72.5 & 87.2 & 84.3 & 88.4 & 52.6& 85.7\\
 DETR-DC5 (R50)~\cite{10.1007/978-3-030-58452-8_13} 
 &41 & 187 & 87.8 & 85.2 & 89.2 & \textbf{54.3}& 86.5\\
 RT-DETR (R50)~\cite{10657220} 
 &42 & 136 & \textbf{91.2} & \underline{87.3} & \underline{92.5} & 53.2& \underline{89.2}\\
 YOLOv5s~\cite{10533615}
 &7.21 & 16.5 & 86.8 & 82.2 & 90.9 & 52.2& 84.4\\
 U-YOLOv7~\cite{YU2023102108}
 &10.91 & 34.2 & 88.4 & 78.2 & 89.2 & 51.4& 83.0\\
 YOLOv8s~\cite{10533618}
 &10.65 & 28.8 & 86.4 & 82.0 & 90.1 & 52.0& 84.1\\
 YOLOv10s~\cite{2024YOLOv10}
 &7.2 & 21.6 & 90.1 & 81.9 & 91.2 & 52.3& 85.8\\
 YOLOv11s~\cite{10533611}
 &9.4 & 21.5 & 87.7 & 81.1 & 88.0 & 52.0& 84.3\\
\rowcolor{gray!20} \textbf{EPA-Net (Ours)} & 9.7 & 35 & \underline{90.2} & \textbf{89.4} & \textbf{92.6} & \underline{53.9} & \textbf{89.8} \\
        \hline
    \end{tabular}
    \label{tab:model_comparison1}
\end{table*}

\subsection{Implementation Details} 
The proposed method is implemented with the PyTorch framework. Following existing work, all reported FLOPs and inference times are calculated using a 640 × 640 image input with batch size = 1. The proposed EPA-Net is optimized for 300 epochs using the SGD~\cite{SGD}, with an initial learning rate of 0.01, and weight decay and momentum set to $5 \times{ 10^{-4}}$ and 0.9, respectively. Training is conducted on a single RTX 4090 GPU.

\begin{table}[t]
\centering
\caption{Comparison of EPA-Net and SoTA methods on \textbf{AquaFishSet} dataset. The best and second-best performances are highlighted in \textbf{bold} and \underline{underlined}, respectively.}
\label{tab:model_comparison2}
\setlength{\tabcolsep}{3pt}
\begin{tabular}{l|c|c|c|c c}
\hline
\textbf{Model} &  \textbf{P}& \textbf{R} & \textbf{mAP$_{50}$ } & \textbf{mAP$_{50-95}$ }\\
\hline
 Faster R-CNN~\cite{7410526}
  & 85.3 & 80.2 & 86.4 & 53.6 \\
  CAM-RCNN~\cite{YI2024127488}
  & 86.0 & 80.8 & 87.0 & 53.4\\
Mask R-CNN~\cite{8237584}
  & 85.8 & 80.6 & 86.9 & 53.2\\
YOLO-Fish~\cite{MUKSIT2022101847}
& 85.5 & 80.4 & 84.9 & 52.2\\
RC-YOLOv5~\cite{Li2023} 
& 87.4 & 81.4 & 86.9 & 54.2\\
RT-DETR (R50)~\cite{10657220} 
& \textbf{92.4} & 89.5 &\underline{92.2} & 55.4\\
 CEH-YOLO~\cite{FENG2024102758}
 &89.0 & 82.8 & 87.2 & 55.2\\
SUR-Net~\cite{9851605} 
& 88.4 & 87.5 & 90.2 & 54.2\\
 YOLOv5n~\cite{10533615}
 & 87.2 & 81.6 & 86.8 & 54.5 \\
  YOLOv8n~\cite{10533618}
  & 88.3 & 83.1 & 88.6 & 55.5\\
 YOLOv10n~\cite{2024YOLOv10}
  & 90.4 & 84.6 & 89.2 & 56.3\\
 YOLOv11n~\cite{10533611}
  & 88.5 & 83.5 & 88.1 & 55.2\\
 YOLOv5s~\cite{10533615}
  & 88.8 & 84.2 & 89.7 & 53.2\\
 U-YOLOv7~\cite{YU2023102108}
 &90.2 & 88.4 & 89.9 & 54.2\\
 YOLOv8s~\cite{10533618}
  & 90.4 & 88.0 & 90.1 & 55.0\\
 YOLOv10s~\cite{2024YOLOv10}
 & 91.2 & 90.4 & \underline{92.2} & \textbf{56.4}\\
 YOLOv11s~\cite{10533611}
  & 90.8 & 89.1 & 90.0 & 54.0\\
        \rowcolor{gray!20} \textbf{EPA-Net (Ours)} & \underline{91.2} & \textbf{90.8} & \textbf{93.6} & \underline{56.2}   \\
\hline
\end{tabular}
\vspace{0.2cm}
\end{table}

\begin{table}[t]
\centering
\caption{Comparison of EPA-Net and SoTA methods on \textbf{UW-BlurredFish } dataset. The best and second-best performances are highlighted in \textbf{bold} and \underline{underlined}, respectively.}
\label{tab:model_comparison3}
\setlength{\tabcolsep}{3pt}
\begin{tabular}{l|c|c|c|c c}
\hline
\textbf{Model} &  \textbf{P}& \textbf{R} & \textbf{mAP$_{50}$} & \textbf{mAP$_{50-95}$ }\\
\hline
  Faster R-CNN~\cite{7410526}
  & 82.4 & 73.6 & 82.4 & 50.4 \\
   CAM-RCNN~\cite{YI2024127488}
  & 83.2 & 77.2 & 83.0 & 50.5\\
Mask R-CNN~\cite{8237584}
  & 83.4 & 75.2 & 82.2 & 50.2\\
YOLO-Fish~\cite{MUKSIT2022101847}
& 82.7 & 78.4 & 82.9 & 50.6\\
RC-YOLOv5~\cite{Li2023} 
& 83.4 & 78.4 & 84.0 & 50.2\\
RT-DETR (R50)~\cite{10657220} 
& \textbf{90.2} & 84.6 &\underline{89.2} & 52.4\\
 CEH-YOLO~\cite{FENG2024102758}
 &88.2 & 82.4 & 85.2 & 51.2\\
SUR-Net~\cite{9851605} 
& 86.4 & 83.5 & 86.2 & 52.2\\
 YOLOv5n~\cite{10533615}
 & 83.2 & 79.6 & 84.8 & 51.2 \\
  YOLOv8n~\cite{10533618}
  & 85.0 & 82.4 & 85.1 & 52.5\\
 YOLOv10n~\cite{2024YOLOv10}
  & 86.5 & \underline{86.6} & 86.6 & 53.3\\
 YOLOv11ns~\cite{10533611}
  & 86.0 & 82.5 & 86.1 & 52.8\\
 YOLOv5s~\cite{10533615}
  & 86.8 & 83.2 & 87.9 & 53.4\\
 U-YOLOv7~\cite{YU2023102108}
 &87.2 & 85.0 & 88.9 & 53.2\\
 YOLOv8s~\cite{10533618}
  & 87.4 & 84.0 & 88.1 & 53.0\\
 YOLOv10s~\cite{2024YOLOv10}
 & 88.1 & 82.9 & 88.2 & \textbf{54.2}\\
 YOLOv11ss~\cite{10533611}
  & 87.7 & 81.1 & 88.0 & 52.0\\
        \rowcolor{gray!20} \textbf{EPA-Net (Ours)} & \underline{89.4} & \textbf{88.5} & \textbf{90.4} & \underline{53.4}   \\
\hline
\end{tabular}
\vspace{0.2cm}
\end{table}

\begin{table}[ht]
\centering
\caption{Disentangling backbone and EPA-FPN, and introducing MS-DDSP Bottleneck,$'\$'$ represents using the MS-DDSP Bottleneck}
\label{tab:backbone_epafpn}
\setlength{\tabcolsep}{3pt}
\begin{tabular}{l|c|c|c}
\hline
  & \textbf{mAP$_{50-95}$} & \textbf{Params} & \textbf{Infer. Time} \\
\hline
ResNet50 + FPN         & 42.5 & 34.0M  & 15 ms \\
CSPDarknet + FPN  & 44.3 & 21.1M  & 12.4 ms \\
CSPDarknet + EPA-FPN & 50.3 & 14.7M  & 8.6 ms \\
\textbf{CSPDarknet + EPA-FPN $\$$} & 53.4 & 14.9M  & 8.6 ms \\
\hline
\end{tabular}
\vspace{0.2cm}
\end{table}

  \begin{figure*}[t]
	    \centering
        \includegraphics[width=0.99\linewidth]{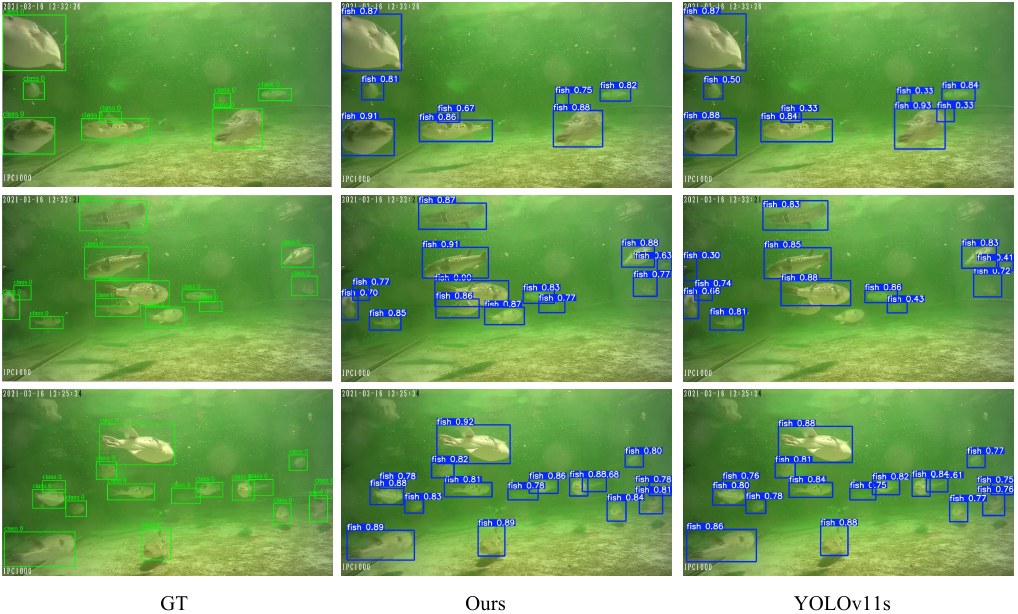}
        \caption{\textbf{EPA-Net architectur --}
        	 It uses CSPDarknet as the backbone network, EPA-FPN as the feature network and forms the neck network together with MS-c2f. The structure of the MS-c2f module is at the bottom of the figure.
            }
       \vskip -0.1in
        \label{fig:Visualization}                                                  
\end{figure*} 

\begin{table}[ht]
\centering
\caption{Comparison of different FPN architectures,All FPNs use CSPDarknet as the backbone.}
\label{tab:fpn_comparison}
\begin{tabular}{l|c|c|c}
\hline
 &  \textbf{mAP$_{50-95}$} & \textbf{Params } & \textbf{Infer. Time} \\
\hline
top-down FPN       & 44.8 & 1.2M  & 12.4 ms  \\
Fully-Connected FPN            & 45.1 & 1.2M  & 13.0 ms  \\
Bi-FPN                    & 47.5 & 1.4M & \textbf{8.6 ms} \\
PANet         & \underline{50.2} & 2.3M & 13.2 ms\\
\textbf{EPA-FPN(ours) } & \textbf{50.3} & 1.6M & \textbf{8.6 ms} \\
\hline
\end{tabular}
\end{table}

\subsection{Comparison with State-of-the-art Methods} 
\subsubsection{\textbf{Comparisons on DeepFish}}
We evaluate the performance of EPANet against 15 SoTA methods on the HRSID dataset. The results in Table \ref{tab:model_comparison1} show that our proposed method achieves state-of-the-art performance on all parameters compared to models with similar parameters and FLOPs (e.g., YOLOv8s and YOLOv11s). Compared to the larger RT-DETR, EPA-Net achieves comparable or even better detection performance with only 23\% of the model size and 26\% of the FLOPs.  It is noteworthy that, in scenarios with very few samples, our EPANet significantly outperforms other methods in terms of Recall and $F_1$ score. This demonstrates the effectiveness of our approach in targeting and utilizing spatial information.

\subsubsection{\textbf{Comparisons on AquaFishSet}}
We compared EPANet with 14 SoTAs on the AquaFishSet dataset, as shown in Table \ref{tab:model_comparison2}. EPANet achieved SoTA with a mAP of 92.6\%, which is +2.5\% and +2.6\% higher than YOLOv8s and YOLOv11s, respectively. It has a lower false positive rate and missed positive rate, which is attributed to the effectiveness of information complementarity and diversion convolution, proving the superiority of our method.
\subsubsection{\textbf{Comparisons on UW-BlurredFish}}
We compared EPANet with 14 other models on the self-built UW-BlurredFish dataset, and the results are shown in Table \ref{tab:model_comparison3}. The results show that in more blurred underwater scenes, the performance of all models has declined, but our EPANet still achieves 90.4\% MAP, which exceeds all other methods. This reflects the robustness of our method in complex environments and further verifies the effectiveness of our method.

\subsection{Ablation Studies and Analysis} 
The comparison results presented in Table \ref{tab:model_comparison1}, Table \ref{tab:model_comparison2}, and Table \ref{tab:model_comparison3}, demonstrate that the proposed EPANet underwater fish detection method is superior to many state-of-the-art methods.
In what follows,  the proposed EPANet is comprehensively analyzed from four aspects to investigate the logic behind its superiority. For simplicity, all accuracy results
here are for UW-BlurredFish validation set.

\subsubsection{\textbf{Contribution of EPANet backbone and each module}}
Since EPANet utilizes a powerful backbone, a new FPN, and a new Bottleneck simultaneously, we aim to understand their respective contributions to enhancing accuracy and efficiency. 
Table \ref{tab:backbone_epafpn} compares the impact of the backbone network and each module. Starting with a RetinaNet detector equipped with a ResNet-50 backbone network and a top-down FPN, we first replace the backbone network with CSPDarknet, which increases the accuracy by approximately 2\% AP, accompanied by a slight reduction in parameters and inference time. Subsequently, by replacing the FPN with our proposed EPA-FPN, we achieve an additional 6\% AP improvement, along with a significant reduction in parameters and inference time. Finally, by introducing the MS-DDSP Bottleneck module, we further enhance the accuracy by 3\% AP with almost no impact on parameters and inference time. These results demonstrate that the EPANet backbone network, EPA-FPN, and MS-DDSP Bottleneck are all crucial to our final model. 

\subsubsection{\textbf{Effect of EPA-FPN}}
Table \ref{tab:fpn_comparison} shows the accuracy and model complexity of the Feature Pyramid Networks with different cross-scale connections listed in Figure \ref{fig:epafpn}. We use the same backbone network and category/box prediction network for all experiments, as well as the same training settings. It can be seen that the traditional top-down FPN is inherently limited by the unidirectional information flow and thus has the lowest accuracy. PANet adopts a strategy of spatial semantic information complementation to make its accuracy slightly higher than Bi-FPN in underwater scenes, but it also requires more parameters. Our EPA-FPN achieves similar accuracy to PANet, but uses fewer parameters and has faster inference speed.

\subsubsection{\textbf{Effect of MS-DDSP Bottleneck}}
Table \ref {tab:bottleneck} shows the accuracy and inference speed of different Bottlenecks under two backbones. We used the same training settings for all experiments. It can be seen that using ResNet50 or CSPDarknet, our MS-DDSP Bottleneck improves the AP by 2\% compared to traditional bottlenecks without significantly affecting the inference speed, fully demonstrating the effectiveness of our module.

\begin{table}[t]
\centering
\caption{Comparison of different FPN architectures, all FPNs use CSPDarknet as the backbone.}
\label{tab:bottleneck}
\begin{tabular}{l|c|c}
\hline
 & \textbf{AP$_{50}$} & \textbf{Infer. Time }   \\
\hline
ResNet50 + Bottleneck        & 37.4 & 15.0 ms     \\
ResNet50 + MS-DDSP Bottleneck            & 39.8 & 15.0 ms    \\
CSPDarknet + Bottleneck                    & 42.2 & \textbf{8.2 ms }\\
\textbf{CSPDarknet +MS-DDSP Bottleneck} & \textbf{44.5} & 8.4 ms\\
\hline
\end{tabular}
\end{table}

\subsection{Visualization} 
The visualization results for UW-BlurredFish data sets are shown in Figure 5. It can be observed that, even in underwater complex scenes, our EPANet successfully detects and recognizes objects, with the network can effectively extract the object details and distinguish them from the background. Notably, EPANet can make full use of the complementarity of spatial semantic information. Even in scenarios of limited resolution, the proposed model maintains a high attention to objects.

\section{Conclusion} 
We analyze the limitations of existing methods in the UFD task, especially their inadequacy in handling low object resolution and high object-background similarity. Based on this, we propose the Efficient Path Aggregation Network (EPANet). EPANet enhances the complementarity of information in feature maps and the diversity of local features through an Efficient Path Aggregation Feature Pyramid Network (EPA-FPN) and a Multi-scale Bottleneck Network (MS-DDSP Bottleneck). Extensive experiments show that, compared with SoTA methods in the UFD task, EPANet achieves state-of-the-art performance with a similar or even smaller number of parameters, which benefits from the efficiency of our model. In the future, we plan to explore extending this approach to the general object detection field and multi-modal tasks.

\section{Acknowledgments} 
This work was supported by the Natural Science Foundation of Liaoning Province under Grant (2024-BS-214), the Basic Scientific Research Project of the Liaoning Provincial Department of Education (JYTQN2023132), the National Natural Science Foundation of China-Liaoning Joint Fund (U23A88A-001), the Key Research and Development Program of Liaoning Province (2023JH2/10200015), and the Key Laboratory of Environment Controlled Aquaculture (Dalian Ocean University), Ministry of Education (202313).

\section*{Data availability}
Data will be made available on request.

\bibliographystyle{unsrt}
\bibliography{cas-dc-template}

\vskip3pt

\bio{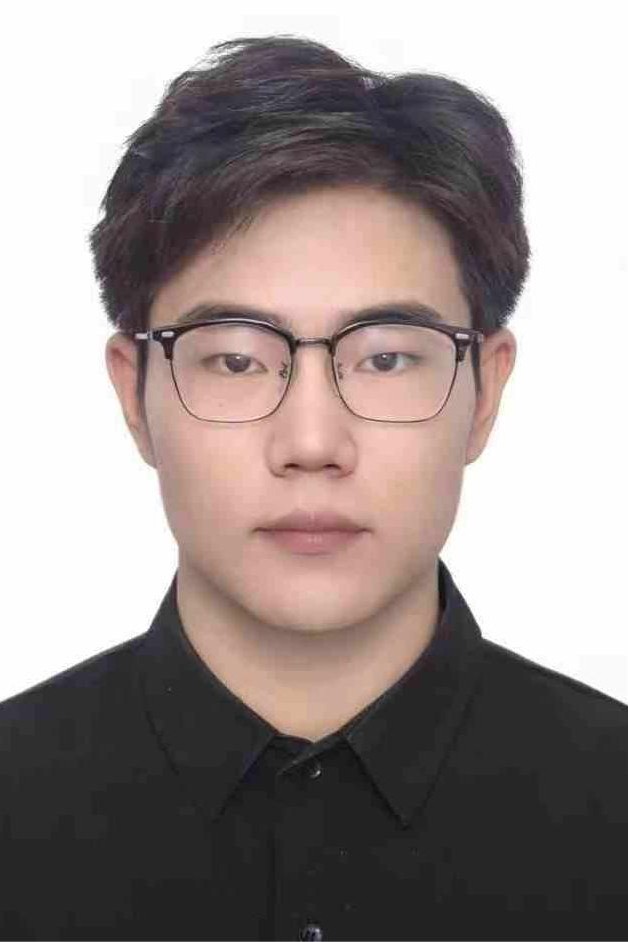}
Jinsong Yang is currently pursuing the B.S. degree in College of Information Engineering at Dalian Ocean University, Dalian, China, since 2022. His research interests include computer vision and deep learning.
\endbio

\vspace{60pt}
\bio{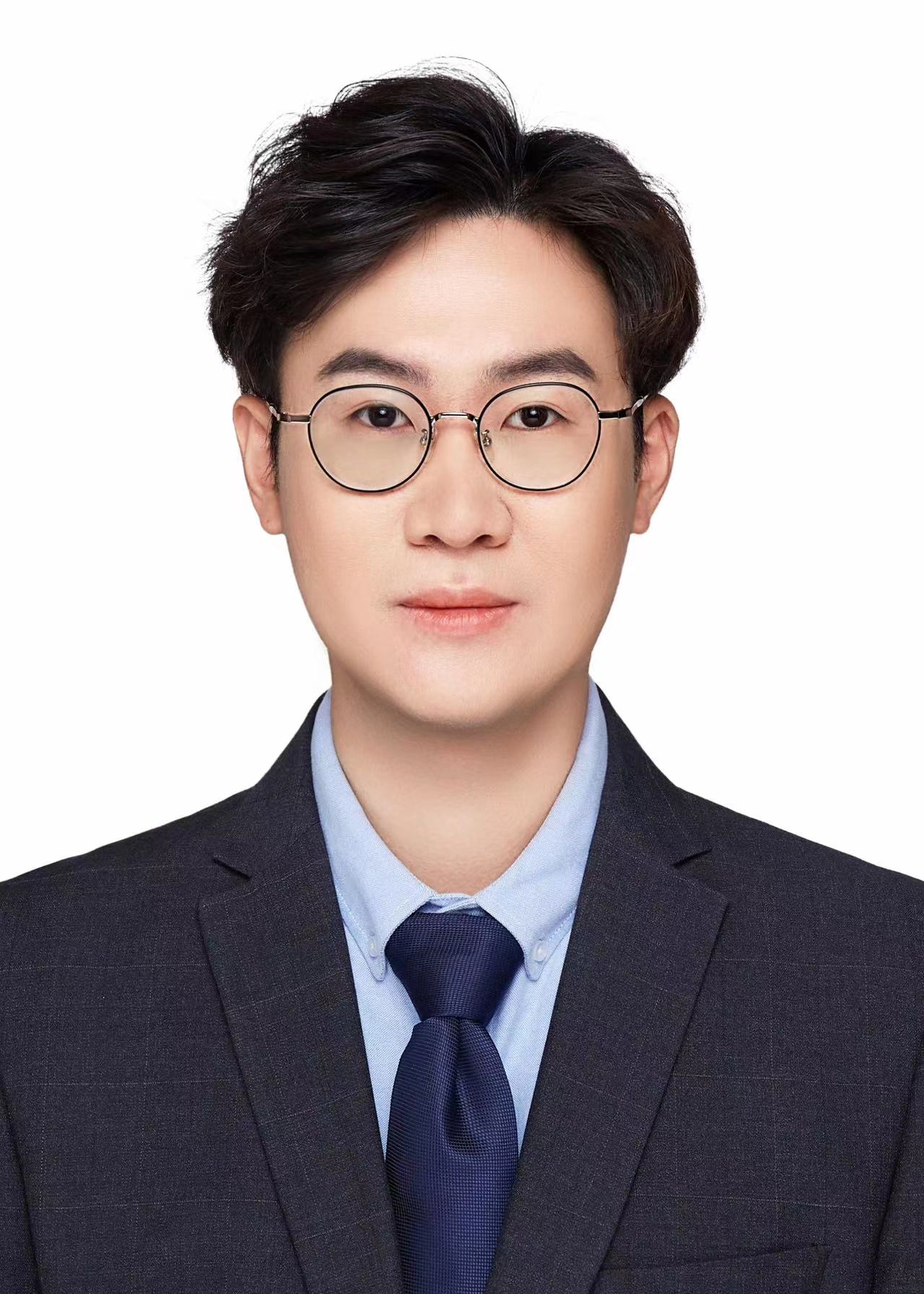}
Zeyuan Hu received the B.S. degree in Automation in 2015, the M.S. and Ph.D. in Information Communication Engineering in Tongmyong University, Busan, South Korea, in 2018 and 2021 respectively. He works as a teacher in College of Information Engineering, Dalian Ocean University, Dalian, China. His main research interests include computer vision, image processing, and deep learning.
\endbio

\vspace{15pt}
\bio{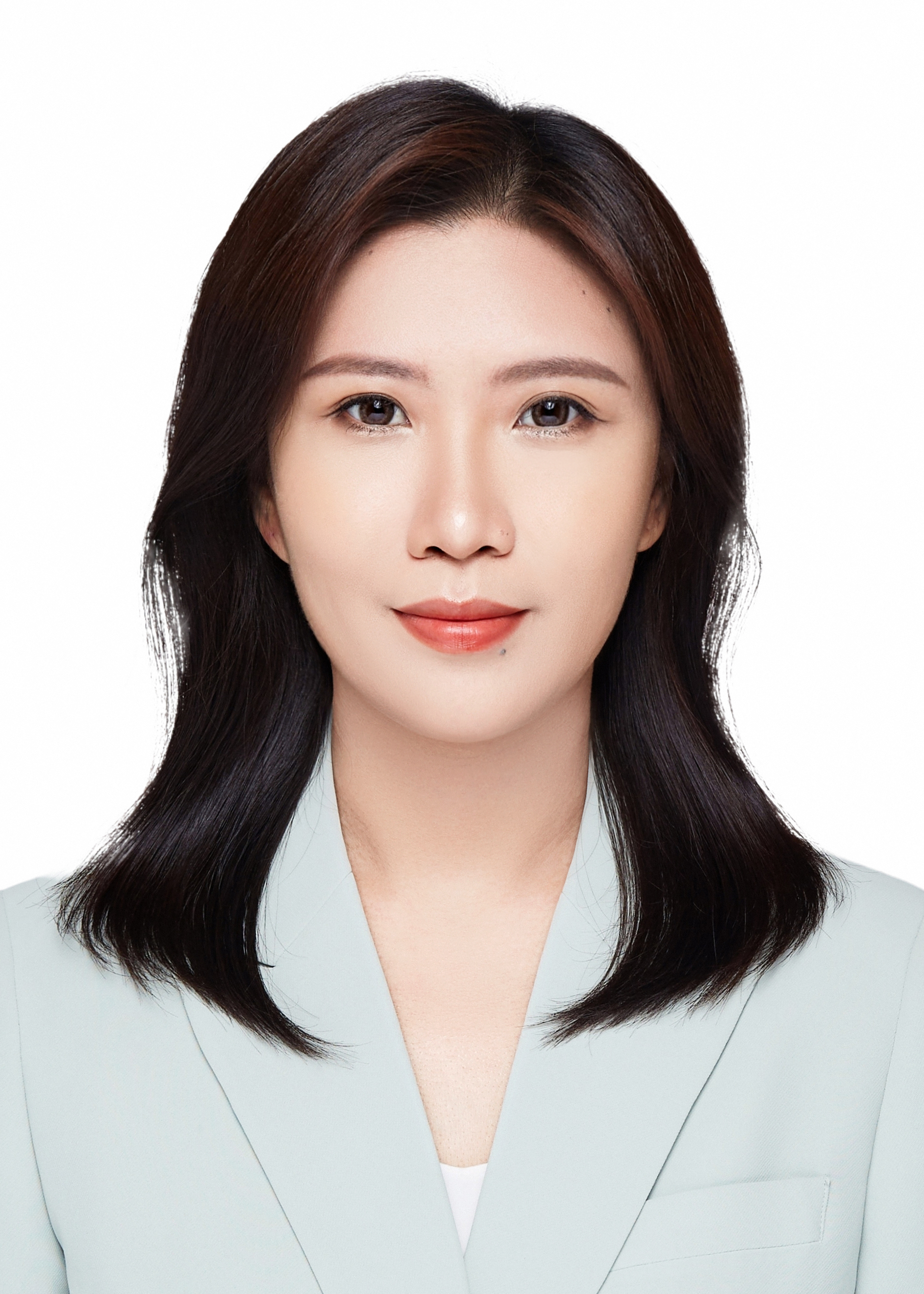}
Yichen Li received the B.A. degree in Economics from University of California, Davis, in 2016, and the M.S. degree in Financial Management from The Australian National University, in 2020. She is currently an Instructor at the School of International Education, Dalian Polytechnic University, Dalian, China. Her research focuses on developing computational models for high-frequency trading, economic forecasting, and quantitative risk management using deep learning and big data analytics. She explores applications of machine learning in financial markets, including algorithmic trading strategies and real-time risk assessment systems.  
\endbio



\end{document}